\documentclass{article}
\usepackage{times}
\usepackage{ml2k}
\usepackage{mlapa}

\usepackage[dvips]{graphicx}
\usepackage{latexsym}
\usepackage{xspace}

\usepackage{amsmath}
\usepackage{amsthm}
\usepackage{amssymb}
\usepackage{algorithm}
\usepackage{algorithmic}

\renewcommand{\(}{\left(}
\renewcommand{\)}{\right)}

\renewcommand{\[}{\left[}
\renewcommand{\]}{\right]}

\renewcommand{\S}{{\cal S}}

\newcommand{\U}{{\cal U}}

\newcommand{\Y}{{\cal Y}}

\DeclareMathOperator{\sign}{sign}

\DeclareMathOperator{\TD}{TD}

\DeclareMathOperator{\GPOMDP}{GPOMDP}
\DeclareMathOperator{\CONJGRAD}{CONJPOMDP}

\DeclareMathOperator{\REINFORCE}{REINFORCE}
\DeclareMathOperator{\MDP}{MDP}
\DeclareMathOperator{\POMDP}{POMDP}

\DeclareMathOperator{\GSEARCH}{GSEARCH}

\DeclareMathOperator{\GRAD}{GRAD}

\newcommand{\pomdpg}{{\mbox{\small$\GPOMDP$}}}

\newcommand{\conjgrad}{{\mbox{\small$\CONJGRAD$}}}

\newcommand{\reinforce}{{\mbox{\small$\REINFORCE$}}}
\newcommand{\mdp}{{\mbox{\small$\MDP$}}}
\newcommand{\mdps}{{\mbox{\small$\MDP$}}s}
\newcommand{\pomdp}{{\mbox{\small$\POMDP$}}}
\newcommand{\pomdps}{{\mbox{\small$\POMDP$}}s}
\newcommand{\gsearch}{{\mbox{\small$\GSEARCH$}}}

\newcommand{\grad}{{\mbox{\small$\GRAD$}}}

\newcommand{\nb}{{\nabla_{\!\!\beta}}}
\newcommand{\Expect}{\mathbold{E}}

\newcommand{\E}{\Expect}
\newcommand{\mathbold}[1]{\mbox{\boldmath $\bf#1$}}
\newcommand{\nablabeta}{{\nabla_{\!\!\beta}}}
\renewcommand{\Re}{{\mathbb R}}
\newcommand{\R}{{\mathbb R}}

\usepackage{color}

\definecolor{darkred}{rgb}{0.7,0.2,0.2}

\definecolor{bgblue}{rgb}{0.04,0.39,0.53}

\newtheorem{theorem}{Theorem}

\newtheorem{assumption}{Assumption}

\begin{document}
\twocolumn[
\icmltitle{Reinforcement Learning in POMDP's via Direct Gradient Ascent}
\icmlauthor{Jonathan Baxter}{Jonathan.Baxter@anu.edu.au}
\icmlauthor{Peter L. Bartlett}{Peter.Bartlett@anu.edu.au}
\icmladdress{Research School of Information Sciences and Engineering, Australian National University, Canberra, Australia.\\}
]

\renewcommand{\floatpagefraction}{0.85}

\begin{abstract}
This paper discusses theoretical and experimental aspects of
gradient-based approaches to the direct optimization of policy
performance in controlled \pomdps. We introduce \pomdpg, a
\reinforce-like algorithm for estimating an approximation to the
gradient of the average reward as a function of the parameters of a
stochastic policy. The algorithm's chief advantages are that it
requires only a single sample path of the underlying Markov chain, it
uses only one free parameter $\beta\in [0,1)$, which has a natural
interpretation in terms of bias-variance trade-off, and it requires no
knowledge of the underlying state. We prove convergence of \pomdpg\ and
show how the gradient estimates produced by \pomdpg\ can be used in a
conjugate-gradient procedure to find local optima of the average
reward. 
\end{abstract}
\vspace*{-5mm}
\section{Introduction}
\label{section:intro}

``Reinforcement learning'' is used to describe the general problem of
training an agent to choose its actions so as to increase its
long-term average reward. The structure of the environment is
typically not known explicitly, so the agent is forced to learn by
interaction with the environment.

In value-function based approaches to reinforcement learning
the agent tries to learn the value of each state, or possibly each
state-action pair. It then chooses the action with the highest value
according to its value function. If the value function is exact then
this approach is known to lead to the optimal policy under quite
general conditions \cite{sutton98,bertsekas96}. However, for many
real-world problems it is intractable to represent the value function
exactly and the agent instead tries to select a good approximation to
the value function from a restricted class (for example, a
neural-network or radial-basis-function class). 
This approach has yielded some remarkable empirical successes in
learning to play games, including checkers
\cite{samuel59}, backgammon \cite{tesauro92,tesauro94}, and chess
\cite{ml_00a}.  Successes outside of the games domain include
job-shop scheduling \cite{zhang95}, and dynamic channel allocation
\cite{singh97}.

While there are many algorithms for training approximate value functions
(see \emcite{bertsekas96,sutton98} for comprehensive treatments), with
varying degrees of convergence guarantees, all these algorithms---and
indeed the approximate value function approach itself---suffer from
a fundamental limitation: the error they seek to minimize does not
guarantee good performance from the resulting policy.
More precisely, there exist infinite horizon Markov Decision Processes
(\mdps) with the following properties. For all $\epsilon > 0$ there is
an approximate value function $V$ with
\begin{equation}
\label{eq:cough1}
\max_{i} |V(i) - V^*(i)| = \epsilon,
\end{equation}
where the $\max$ is over all states $i$ and $V^*(i)$ is the true value
of state $i$ under the optimal policy. However, the greedy policy
based on this approximate value function has expected discounted
reward
\begin{equation}
\label{eq:cough2}
\eta = \eta^* -\frac{2 \alpha \epsilon}{1-\alpha},
\end{equation}
where $\eta^*$ is the expected discounted reward of the optimal
policy and $\alpha \in [0,1)$ is the discount factor
\cite{bertsekas96,singh94}.  Thus, even accurate approximations
to the optimal value function can generate bad greedy policies if $\alpha$
is close to $1$.

Because Equation~\eqref{eq:cough2} also defines the {\em worst} expected
discounted reward of any greedy policy derived from an approximate
value function satisfying \eqref{eq:cough1}, it has sometimes been
used as a {\em motivation} for using approximate value function
techniques. However, there are two objections to this. The first is
that most existing algorithms for training approximate value functions
do not minimize the maximum norm between $V$ and $V^*$, but typically
some $\ell_2$ (mean-squared) norm. Secondly, even if these algorithms did minimize
the maximum norm directly, the smallest achievable error $\epsilon$
will be so large in many problems of practical interest that the bound
\eqref{eq:cough2} will be useless. Put another way, if we can choose a $V$
to make $\epsilon$ arbitrarily small in \eqref{eq:cough1}, then we are
not really in an approximate value function setting in the first place.

The fact that the class of approximate value functions does not contain
a value function with small approximation error does not preclude it
from containing a value function whose greedy policy approaches or
even equals the performance of the optimal policy. All that matters
for the performance of the greedy policy is the relative ordering
the approximate value function assigns to the successor states
in each state. This motivates an alternative approach: instead of
seeking to minimize \eqref{eq:cough1} or an $\ell_2$ variant, one
should minimize some form of relative error between state values
\cite{baird93,bertsekas97,tr_std_99}. While this idea is promising,
the approach we take in this paper is even more direct: search for a
policy maximizing the expected discounted reward directly.

We can view the average reward \eqref{eq:cough2} as a function
$\eta(\theta)$ of $\theta\in\R^K$, where $\theta$ are
the parameters of $V$. Provided the dependence of $\eta$ on $\theta$ 
is differentiable, we can compute 
$
\nabla \eta(\theta)
$
and then take a small step in the gradient direction in order 
to increase the average reward. Under general assumptions, such an
approach will converge to a local maximum of the average reward
$\eta$. In general, a {\em greedy} policy based
on $V(\theta)$ will give a non-differentiable $\eta(\theta)$. Thus,
in this paper we consider stochastic policies that generate
distributions over actions rather than a deterministic action. 

We cast our results and algorithms in the formal framework of
Partially Observable Markov Decision Processes (\pomdps). The
advantage of this framework is that it models uncertainty both in the
state-transitions of an agent and in the observations the agent
receives.  The first contribution of this paper is \pomdpg, an
algorithm for computing an approximation, $\nb \eta(\theta)$, to the
true gradient $\nabla \eta(\theta)$, from a single sample path of a
\pomdp.  The algorithm requires storage of only $2K$ real
numbers---where $K$ is the number of parameters in the policy---and
needs no knowledge of the underlying state. The accuracy of the
approximation $\nb \eta(\theta)$ is controlled by a parameter $\beta
\in [0,1)$ (a discount factor) and, in particular,
the accuracy is controlled by the relationship between $\beta$ and the
mixing time of the Markov chain underlying the \pomdp\ (loosely
speaking the mixing time is the time needed to approach stationarity
from an arbitrary starting state).  $\nb\eta(\theta)$ has the property
that $\lim_{\beta\to 1}\nb
\eta(\theta) =
\nabla\eta(\theta)$. However, the trade-off preventing the setting of
$\beta$ arbitrarily close to $1$ is that the variance of the
algorithm's estimates increase as $\beta$ approaches $1$.  We prove
convergence with probability 1 of \pomdpg. 

The second contribution of this paper is \conjgrad, a
conjugate-gradient based optimization procedure that utilizes the
estimates generated by \pomdpg. The key difficulty in performing
greedy stochastic optimization is knowing when to terminate a line
search, since noisy estimates make it very difficult to locate a
maximum. We solve this problem in \conjgrad\ by using gradient
estimates to bracket the maximum, rather than value estimates.

Finally, we present the results of an experiment which illustrates the
key ideas of the paper. All proofs have been omitted due to space
constraints. 
\section{Related Work}
The gradient approach to reinforcement learning was pioneered by
\emcite{williams92}, who introduced the \reinforce\ algorithm
for estimating the gradient in {\em episodic} tasks, for which there
is an identified recurrent state $i^*$, and the algorithm returns a
gradient estimate each time $i^*$ is entered.  Williams showed that
the expected value of this estimate is the gradient direction, in the
case that the number of steps between visits to $i^*$ is a
constant. Other formulae for the performance gradient of a Markov Decision
Process that rely on the existence of a recurrent state have been
given in \emcite{glynn86,cao97,cao98,fu94}, and for \pomdps\ in
\emcite{jaakola95}. Williams' algorithm was generalized to the
infinite-horizon setting in \emcite{kimura97} and to more general reward
structures in \emcite{marbach98}.  The ``VAPS'' algorithm described in
\emcite{baird98} showed how the algorithms in \emcite{marbach98} could be
interpreted as combining both value-function and policy-gradient
approaches.  VAPS also relies on the existence of recurrent states to
guarantee convergence. VAPS can be viewed broadly under
the banner of ``Actor-Critic'' algorithms
\cite{BarSutAnd83},  which have more recently been investigated in
\emcite{singh95,sutton99,konda99}. 

Policy-gradient algorithms for which convergence results have been
proved all rely on the existence of an identifiable recurrent state,
and the variance of these algorithms is related to the time between
visits to the recurrent state.  Although the assumptions we make in
this paper about the \pomdp\ ensure that every state is recurrent, we
would expect that as the size of the state space increases, there will
be a corresponding increase in the expected time between visits to the
identified recurrent state. Furthermore, the time between visits
depends on the parameters of the policy, and states that are
frequently visited for the initial value of the parameters may become
very rare as performance improves.  In addition, in an arbitrary
\pomdp\ it may be difficult to estimate the underlying states, and
therefore to determine when the gradient estimate should be
updated. Thus, a key advantage of \pomdpg\ is that its running time is
not bounded by the requirement to visit recurrent states. Instead, it
is bounded by the ``mixing'' time of the \pomdp\ (loosely, the time
needed to approach stationarity), which is always shorter than
recurrence time and often substantially so.

Approximate algorithms for computing the gradient were also given in
\emcite{marbach98,marbachthesis98}, one that sought to solve the aforementioned
recurrence problem by demanding only recurrence to one of a set of
recurrent states, and another that abandoned recurrence and used
discounting, which is closer in spirit to our algorithm.

\section{The Mathematical Framework}
\label{section:rl}
Our setting is that of an agent taking actions in an environment
according to a parameterized policy. The agent seeks to adjust its
parameters in order to maximize the long-term average reward. We
pursue a local approach: get the agent to compute the gradient of the
average reward with respect to its parameters, and then adjust the
parameters in the gradient direction. Formally, the most natural
setting for this problem is that of Partially Observable Markov
Decision Processes or \pomdps.  For ease of exposition we 
consider finite \pomdps. General results in the continuous case
can be found in \emcite{jair_01a}. 

Specifically, assume that there are $n$ states $\S=\{1,\dots, n\}$ of
the world (including the agent's states), $N$ controls $\U = \{1,
\dots,N\}$ and $M$ observations $\Y =\{1,\dots, M\}$. For each state
$i\in\S$ there is a corresponding reward $r(i)$. Each $u\in \U$
determines a stochastic matrix $P(u) = [p_{ij}(u)]$ where $p_{ij}(u)$
is the probability of making a transition from state $i$ to state $j$
give control $u$.  For each state $i\in\S$, an observation $y\in\Y$ is
generated independently according to a probability distribution
$\nu(i)$ over observations in $\Y$.  We denote the probability of
observation $y$ by $\nu_y(i)$.  A {\em randomized policy} is simply a
function $\mu$ mapping observations $y\in\Y$ into probability
distributions over the controls $\U$.  That is, for each observation
$y$, $\mu(y)$ is a distribution over the controls in $\U$.  Denote the
probability under $\mu$ of control $u$ given observation $y$ by
$\mu_u(y)$. In general, to perform optimally, the policy 
has to be a function of the entire history of observations, but this
can be achieved by concatenating observations and treating
the vector of observations as input to the policy. One could also
consider policies that have memory, such as parameterized finite
automata, but that is the subject of some of our ongoing research and
is beyond the scope of the present paper. 

To each randomized policy $\mu(\cdot)$ and observation distribution
$\nu(\cdot)$, there corresponds a Markov chain in which state transitions
are generated by first selecting an observation $y$ in state $i$
according to the distribution $\nu(i)$, then selecting a control $u$
according to the distribution $\mu(y)$, and then generating a transition
to state $j$ according to the probability $p_{ij}(u)$. To parameterize
these chains we parameterize the policies, so that $\mu$ now becomes
a function $\mu(\theta, y)$ of a set of parameters $\theta\in\R^K$ as
well as the observation $y$. The Markov chain corresponding to $\theta$
has state transition matrix $P(\theta) = [p_{ij}(\theta)]$ given by
$
p_{ij}(\theta) = \E_{y\sim \nu(i)} \E_{u\sim \mu(\theta, y)} p_{ij}(u). 
$
Throughout, we assume that these Markov chains satisfy the following
assumption:
\begin{assumption}
\label{ass:station}
\sloppy
Each $P(\theta)$ has a unique stationary distribution $\pi(\theta)
:= \[\pi(\theta, 1), \dots, \pi(\theta,n)\]'$ satisfying the {\em
balance equations}
$\pi'(\theta) P(\theta) = \pi'(\theta)$
(throughout $\pi'$ denotes the transpose of $\pi$).
The magnitudes of the rewards, $|r(i)|$, are uniformly bounded by
$R < \infty$ for all states $i$.
\end{assumption}
Our goal is to find a $\theta \in\R^K$ maximizing the {\em long-term average
reward}:
$$
\eta(\theta) := \lim_{T\rightarrow\infty} \frac1T\E_\theta \[\sum_{t=1}^T r(i_t)\].
$$
where $\E_\theta$ denotes the expectation over all sequences $i_0, i_1,
\dots,$ with transitions generated according to $P(\theta)$. Under our 
assumptions, $\eta(\theta)$ is independent of the starting state $i_0$ 
and is equal to:
\begin{equation}
\eta(\theta) = \sum_{i=1}^n \pi(\theta,i) r(i) = \pi'(\theta) r,
\label{eq:avgreward}
\end{equation}
where $r = \[r(1), \dots, r(n)\]'$ \cite{bertsekas95}.

In contrast to the average reward, many approximate value-function based
algorithms such as $\TD(\lambda)$ seek to optimize with respect to the
expected {\em discounted} reward, where the latter is defined by  
$
\label{eq:Jdef}
\eta_\alpha(\theta) := \sum_i \pi_i J_\alpha(\theta,i) 
= \sum_i \pi_i \E_\theta \[\sum_{t=0}^\infty \alpha^t r(i_t) | i_0 =
i\] 
$
(here $\alpha\in [0,1)$ is the discount factor).  In this case the
discounted value of state $i$, $J_\alpha(\theta,i)$, does depend on
the starting state $i$.  A curious fact about the present setting is
that optimizing the long-term average reward is the same as optimizing
expected discounted reward, since 
$\eta_\alpha(\theta) = \eta(\theta) / (1- \alpha)$ \cite[Fact
7]{singh94a}. So without loss of generality we can consider just
average reward.

\section{Gradient Ascent on $\mathbold{\eta(\theta)}$}
\label{sec:gradascent}

The approach taken to optimization of $\eta(\theta)$ in
this paper is {\em gradient ascent}.  That is, repeatedly compute
$\nabla\eta(\theta)$ with respect to the parameters $\theta$,
and then take a step in the uphill direction: $\theta \leftarrow \theta +
\gamma \nabla\eta(\theta)$, for some suitable step-size $\gamma$.

A straightforward calculation shows that 
\begin{equation}
\label{eq:gradeta}
\nabla \eta = \nabla\pi' r = \pi'\nabla P\[I - P + e \pi'\]^{-1} r,
\end{equation}
where $e\pi'$ is the square matrix with each row equal to the
stationary distribution $\pi'$ \cite{jair_01a}. Note that
\eqref{eq:gradeta} should be read as $K$ equations, one for each of
the partial derivatives $\partial/\partial\theta_i$.
For \pomdps\ with a sufficiently small number of states (and known
transition probabilities and observation probabilities), \eqref{eq:gradeta}
could be solved exactly to yield the precise gradient direction.  This 
may be an interesting avenue for further investigation since
\pomdps\ are generally intractable even for small numbers of states
\cite{papadimitriou87}. However, in general the transition and
observation probabilities will be unknown, and the state-space too
large for the matrix inversion in the right-hand-side of
\eqref{eq:gradeta} to be feasible. 
Thus, for many problems of practical interest, \eqref{eq:gradeta} will be
intractable and we will need to find some other way of computing the
gradient. One approximate technique for doing this is presented in the
next section.

\section{Approximating the Gradient $\mathbold{\nabla \eta(\theta)}$}
\label{sec:approx}

In this section, we show that the gradient can be split into two
components, one of which becomes negligible as a discount factor
$\beta$ approaches $1$.

Recall the definition of the discounted value of
state $i$, $J_\beta(\theta,i)$. Here $\beta\in[0,1)$ is the discount
factor. Write $J_\beta(\theta) = \[J_\beta(\theta, 1), \dots,
J_\beta(\theta, n)\]'$ or simply $J_\beta = \[J_\beta(1), \dots,
J_\beta(n)\]'$ when the dependence on $\theta$ is obvious.
\begin{theorem}
\label{theorem:factor}
For all $\theta\in\R^K$ and $\beta \in [0,1)$, 
\begin{equation}
\label{eq:factor}
\nabla\eta = (1-\beta)\nabla\pi'J_\beta +
\beta\pi' \nabla P J_\beta.
\end{equation}
\end{theorem}
We shall see in the next section that the second term
in~\eqref{eq:factor} can be estimated from a single sample path of the
\pomdp.
The following theorem shows that the first term in~\eqref{eq:factor}
becomes negligible as $\beta$ approaches $1$.  Notice that this is not
immediate from Theorem~\ref{theorem:factor}, since $J_\beta$ can become
arbitrarily large in the limit $\beta\to 1$.

\begin{theorem}
\label{theorem:approx}
For all $\theta\in\R^K$,
\begin{equation}
\label{eq:approx}
\nabla \eta = \lim_{\beta\to 1}\nb \eta,
\end{equation}
where 
\begin{equation}
\label{eq:approxgrad}
\nb \eta := \pi'\nabla P J_\beta.
\end{equation}
\end{theorem}
Theorem~\ref{theorem:approx} shows that $\nb\eta$ is a good
approximation to the gradient as $\beta$ approaches $1$, but it turns
out that values of $\beta$ very close to $1$ lead to large variance in
the estimates of $\nb\eta$ that we describe in the next section (that
is, the estimates produced by $\pomdpg$).  However, the following
theorem shows that $1-\beta$ need not be too small, provided the
Markov chain corresponding to $P(\theta)$ has a short {\em mixing
time}. From any initial state, the distribution over states of a
Markov chain converges to the stationary distribution, provided
Assumption~\ref{ass:station} about the existence and uniqueness of the
stationary distribution is satisfied~\cite[Theorem~15.8.1, p.~552]{lancaster85}.

To precisely quantify mixing time, let $\|p - q\|_1$ denote the
usual $\ell_1$ distance on distributions 
$p = (p_1, \dots, p_n), q= (q_1,\dots, q_n)$: 
$
\|q - p\|_1 = \sum_{i=1}^n |q_i - p_i|.
$
Let $p^t(i)$ denote the distribution over the states of the Markov
chain at time $t$, starting from state $i$. Define $d(t)$ by 
$
d(t) := \max_{i,j\in\S} \|p^t(i) - p^t(j)\|_1.
$
Note that $d(t)$ is a function of the parameters $\theta$ via the
transition matrix $P(\theta)$, and since the state distribution
converges to $\pi(\theta)$ for each $\theta$, $d(t)$ must converge to 
zero. Finally, define the {\em mixing time} $\tau^*(\theta)$ of the
Markov chain by: 
\begin{equation} 
\label{eq:mixtime}
\tau^*(\theta) := \min\left\{t\colon d(t) \leq e^{-1}\right\}.
\end{equation}
\begin{theorem}
\label{thm:mix}
There exists a universal constant $C = C(B,R,n)$ such that 
for all $\beta\in[0,1)$ and $\theta\in\Re^k$, 
\begin{equation}
\label{eq:bias}
\left\| \nabla \eta(\theta) - \nablabeta\eta(\theta) \right\|
	\le C \tau^*(\theta)(1-\beta),
\end{equation}
where $B$ and $R$ are the bounds on $|\nabla\mu/\mu|$ and the
rewards respectively, $n$ is the number of states in the Markov
chain, and the norm $\|\cdot\|$ is the usual two-norm. 
\end{theorem}
Theorem \ref{thm:mix} shows that provided $1/(1-\beta)$ is large
compared with the mixing time $\tau^*(\theta)$,
$\nablabeta\eta(\theta)$ will be a good approximation to
$\nabla\eta(\theta)$. Of course, in general the mixing time
$\tau^*(\theta)$ will be unknown, but the purpose of Theorem
\ref{thm:mix} is not so much to provide a prescription for choosing
$\beta$, but to enhance our understanding of the role $\beta$ plays in
the accuracy of the approximation $\nb\eta(\theta)$. 

\section{Estimating $\mathbold{\nablabeta\eta(\theta)}$}
Having shown that $\nb\eta(\theta)$ can be made a sufficiently
accurate approximation to $\nabla\eta(\theta)$ by choosing the discount
factor $\beta$ judiciously in relation to the mixing time of the
underlying Markov chain, we now describe \pomdpg, an algorithm for
estimating $\nablabeta\eta(\theta)$ from a single sample path of the
$\pomdp$. To understand the algorithm, recall that the \pomdp\
iterates as follows: at time step $t$ the environment is in some state
which we denote by $i_t$. An observation $y_t$ is generated according to
the distribution $\nu(i_t)$. The agent generates a control $u_t$
according to the distribution given by its policy $\mu(\theta,
y_t)$. Finally, the environment makes a transition to a new state
$i_{t+1}$ according to the probability $p_{i_ti_{t+1}}(u_t)$. \pomdpg\
is described in Algorithm \ref{algorithm:pgradmdp}. Note that the
update for $\Delta_t$ is recursively computing the average of $r(i_t)
z_t$. 
\begin{algorithm}
\caption{The \pomdpg\ algorithm.}
\label{algorithm:pgradmdp}
\begin{algorithmic}[1]
\STATE {\bf Given: }
\begin{itemize}
\item Parameterized class of randomized policies $\left\{\mu(\theta, \cdot):
\theta \in \R^K\right\}$ satisfying Assumption~\ref{ass:mubound}.
\item Partially observable Markov decision process which when controlled
by the randomized policies $\mu(\theta, \cdot)$ corresponds to a
parameterized class of Markov chains satisfying Assumption~\ref{ass:station}.
\item $\beta \in [0,1)$.
\item Arbitrary (unknown) starting state $i_0$.
\item Observation sequence $y_0, y_1, \dots$  generated by the \pomdp\
with controls $u_0, u_1, \dots$ generated randomly according
to $\mu(\theta, y_t)$. 
\item Bounded reward sequence $r(i_0),r(i_1),\dots$, 
where $i_0,i_1,\dots$ is the (hidden) sequence of states of the Markov decision
process.
\end{itemize}
\STATE Set $z_0 = 0$ and $\Delta_0 = 0$ ($z_0, \Delta_0 \in\R^K$). 
\FOR{each observation $y_t$, control $u_t$, and subsequent reward $r(i_{t+1})$}
\STATE $z_{t+1} = \beta z_t + \frac{\nabla \mu_{u_t}(\theta, y_t)}{\mu_{u_t}(\theta, y_t)}$
\STATE $\Delta_{t+1} = \Delta_t + \frac1{t+1}\[r(i_{t+1}) z_{t+1} - \Delta_t\]$
\ENDFOR
\end{algorithmic}
\end{algorithm}
We now show that the estimate $\Delta_t$ produced by \pomdpg\ at time
step $t$ converges to $\nb\eta(\theta)$ as the running time $t$
approaches infinity.  For this we need one more assumption:
\begin{assumption}
The derivatives,
$
\frac{\partial \mu_u(\theta, y)}{\partial \theta_k}
$
exist for all $u\in \U$, $y\in \Y$ and $\theta \in \R^K$. 
\label{ass:mubound}
The ratios 
$$
\[\frac{\left|\frac{\partial \mu_u(\theta, y)}{\partial
\theta_k}\right|}{\mu_u(\theta, y)}\]_{y = 1\dots M; u = 1\dots N; k=1\dots K} 
$$
are uniformly bounded by $B < \infty$ for all $\theta\in \R^K$.
\end{assumption}
This assumption should not be surprising since $\nabla\mu/\mu$ appears
in the update of $z_t$ in \pomdpg. Since $\mu$ appears in the
denominator, we require that if the probability goes to zero for some
$\theta$, then so too must the gradient (and at at least the same
rate).  
\begin{theorem}
\label{theorem:pgradmdp}
Under Assumptions~\ref{ass:station} and \ref{ass:mubound},
Algorithm \ref{algorithm:pgradmdp} starting from any initial state $i_0$ will
generate a sequence $\Delta_0, \Delta_1, \dots, \Delta_t, \dots$
satisfying
\begin{equation}
\label{pconverge}
\lim_{t\to\infty} \Delta_t = \nb\eta \quad\text{\rm{w.p.1}}.
\end{equation}
\end{theorem}
Theorem \ref{theorem:pgradmdp} provides a characterization of the
limiting behavior of \pomdpg. In fact, we also have a result
characterizing the finite time behavior of \pomdpg. Loosely speaking,
provided
\begin{equation}
\label{eq:var}
t > \Omega\(\frac{\tau^*(\theta)}{\epsilon^2(1-\beta)^2}\), 
\end{equation}
then $\|\Delta_t - \nb\eta(\theta)\|_\infty < \epsilon$ with high
probability (see
\emcite{colt_00} for a more precise statement). Comparing
\eqref{eq:var}  and \eqref{eq:bias}, we can see the bias/variance
tradeoff inherent in the choice of $\beta$: equation \eqref{eq:bias}
tells us that to reduce the bias in the estimate $\nb\eta(\theta)$ we
must set $\beta$ close to 1, while \eqref{eq:var} indicates that to
reduce the variance in the estimates of $\nb\eta(\theta)$ produced by
\pomdpg\ at time $t$, we should set $\beta$ as close to $0$ as
possible. 

\section{Stochastic Gradient Ascent Algorithms}
One technique for optimizing \pomdps\ using \pomdpg\ would be to
repeatedly compute $\Delta_T(\theta)$ (the estimate produced by
\pomdpg\ after $T$ iterations with policy parameters $\theta$), and
then update the parameters by $\theta \leftarrow \theta + \gamma
\Delta_T(\theta)$ for a suitable step-size $\gamma$. However, since
the number of iterations $T$ needed to ensure low variance in the
estimates $\Delta_T$ can be quite large, we would like to make more
efficient usage of the estimates by searching for a maximum in the
direction $\Delta_T$. 
\begin{algorithm}
\caption{$\quad\conjgrad(\grad, \theta, s_0, \epsilon) \to \Re^K$}
\label{algorithm:conjgrad}
{\small
\begin{algorithmic}[1]
\STATE {\bf Given: }
\begin{itemize}
\item $\grad\colon \R^K\to \R^K$: an
estimate of the gradient of the objective function to be maximized. 
\item Starting parameters $\theta \in \R^K$
\item Initial step size $s_0 > 0$.
\item Gradient resolution $\epsilon$. 
\end{itemize}
\STATE $g = h = \grad(\theta)$
\WHILE{$\|g\|^2 \ge \epsilon$}
	\STATE $\gsearch(\grad, \theta, h, s_0, \epsilon)$
	\STATE $\Delta = \grad(\theta)$
	\STATE $\gamma = (\Delta - g) \cdot \Delta / \|g\|^2$
	\STATE $h = \Delta + \gamma h$
	\IF{$h \cdot \Delta < 0$}
	\STATE $h = \Delta$
	\ENDIF
	\STATE $g = \Delta$
\ENDWHILE
\STATE return $\theta$
\end{algorithmic}
}
\end{algorithm}
\begin{algorithm}
\caption{$\quad\gsearch(\grad, \theta_0, \theta^*, s,
\epsilon)\to\Re^K$}
\label{algorithm:linesearch}
{\small
\begin{algorithmic}[1]
\STATE {\bf Given: }
\begin{itemize}
\item $\grad\colon \R^K\to \R^K$: gradient estimate.
\item Starting parameters $\theta_0 \in \R^K$.
\item Search direction $\theta^*\in \R^K$ with $\grad(\theta_0)
\cdot \theta^* > 0$.
\item Initial step size $s > 0$.
\item Inner product resolution $\epsilon >= 0$. 
\end{itemize}
\STATE $\theta = \theta_0 + s \theta^* $
\STATE $\Delta =  \grad(\theta)$
\IF{$\Delta \cdot \theta^* < 0$}
	\STATE{Step back to bracket the maximum:}
	\REPEAT
		\STATE $s_+ = s$, $p_+ = \Delta \cdot \theta^*$, $s = s/2$
		\STATE $\theta = \theta_0 + s \theta^*$
		\STATE $\Delta = \grad(\theta)$
	\UNTIL{$\Delta \cdot \theta^* > -\epsilon$}
	\STATE $s_- = s$
	\STATE $p_- = \Delta \cdot \theta^*$
\ELSE
	\STATE{Step forward to bracket the maximum:}
	\REPEAT
		\STATE $s_- = s$, $p_- = \Delta \cdot \theta^*$, $s = 2 s$
		\STATE $\theta = \theta_0 + s \theta^*$
		\STATE $\Delta = \grad(\theta)$
	\UNTIL{$\Delta \cdot \theta^* < \epsilon$}
	\STATE $s_+ = s$
	\STATE $p_+ = \Delta \cdot \theta^*$
\ENDIF
\IF{$p_- > 0$ and $p_+<0$}
	\STATE $s = \frac{s_- p_+ - s_+ p_-}{p_+ - p_-}$
\ELSE 
	\STATE $s = \frac{s_- + s_+}{2}$
\ENDIF
\STATE return $\theta_0 + s \theta^*$
\end{algorithmic}
}\
\end{algorithm}\vspace{-3mm}
\sloppy
\conjgrad, described in Algorithm~\ref{algorithm:conjgrad}, is
a version of the Polak-Ribiere conjugate-gradient algorithm
\cite[\S5.5.2, for example]{fine99} that is designed to operate using only
noisy (and possibly) biased estimates of the gradient of the objective
function (for example, the estimates $\Delta_T$ provided by \pomdpg).
The argument $s_0$ to \conjgrad\ provides an initial step-size for
\gsearch.  When $\|\grad(\theta)\|^2$ falls below the argument $\epsilon$,
\conjgrad\ terminates.

The linesearch algorithm \gsearch\
(Algorithm~\ref{algorithm:linesearch}) uses only gradient information
to bracket the maximum, and then uses quadratic interpolation to jump
to the maximum.  To bracket the maximum in the direction $\theta^*$
from $\theta$,
\gsearch\ finds two points $\theta_1$ and $\theta_2$ in that direction
such that $\grad(\theta_1) \cdot \theta^* > 0$ and $\grad(\theta_2)
\cdot \theta^* < 0$.  This approach is far more robust than the use
of function values.  Even if the estimates $\grad(\theta)$ are noisy,
the variance of $\sign\left[\grad(\theta_1) \cdot \theta^*\right]$
is independent of the distance between $\theta_1$ and $\theta_2$.
(In contrast, the variance of a comparison of function values at two
points increases as the points get closer together.) The disadvantage
is that it is not possible to detect extreme overshooting of the maximum
using only gradient estimates. However, with careful control of the line
search we did not find this to be a problem.


\conjgrad\ operates by iteratively choosing ``uphill'' directions and
then searching for a local maximum in the chosen direction. In the
rest of the paper, we assume that the \grad\ argument to \conjgrad\ is
\pomdpg. 

\section{Experiments} 
Due to space constraints we only have room to consider one experiment, 
and we have chosen a ``toy'' problem so that we can illustrate all the 
key ideas from the rest of the paper. Experiments closer to ``reality''
are discussed in \emcite{jair_01b}.

Consider a three-state \mdp, in each state of which there is a choice
of two actions $a_1$ and $a_2$.  Table~\ref{tab:threestate} shows the
transition probabilities as a function of the states and actions. Each
state $x$ has an associated two-dimensional feature vector $\phi(x) =
(\phi_1(x), \phi_2(x))$, with values of $(12/18, 6/18), (6/18,12/18),
(5/18,5/18)$ for each of $A$, $B$ and $C$ respectively. The reward is
$1$ in state $C$ and $0$ for the other two states.  
Clearly, the optimal policy is to always select the
action that leads to state $C$ with the highest probability, which
from Table \ref{tab:threestate} means always selecting action $a_2$.
\begin{table}
\begin{center}
\begin{tabular}{|c|c|c|c|c|}
\hline
Origin &  &
\multicolumn{3}{|c|}{Destination State Probabilities}\\
State & Action  & $A$    & $B$    &    $C$ \\
\hline
$A$       &   $a1$   & ~~~~~0.0~~~~  & ~~~~0.8~~~~ & 0.2 \\
$A$       &   $a2$   & ~~~~~0.0~~~~  & ~~~~0.2~~~~ & 0.8 \\ \hline
$B$       &   $a1$   & ~~~~~0.8~~~~  & ~~~~0.0~~~~ & 0.2 \\
$B$       &   $a2$   & ~~~~~0.2~~~~  & ~~~~0.0~~~~ & 0.8 \\ \hline
$C$       &   $a1$   & ~~~~~0.0~~~~  & ~~~~0.8~~~~ & 0.2  \\
$C$       &   $a2$   & ~~~~~0.0~~~~  & ~~~~0.2~~~~ & 0.8 \\
\hline
\end{tabular}
\caption{Transition probabilities of the three-state
MDP\label{tab:threestate}}\vspace*{-5mm}
\end{center}
\end{table}
This rather odd choice of feature vectors for the states ensures that
a value function linear in those features and trained using
$\TD(1)$---while observing the optimal policy---will implement a
suboptimal one-step greedy lookahead policy itself
\cite{tr_std_99}. Thus, in contrast to the gradient based approach,
for this system, $\TD(1)$ training a linear value function is
guaranteed to produce a worse policy if it starts out observing the
optimal policy.
\vspace{-3mm}
\subsection{Training a Controller}
\vspace{-2mm}
Our goal is to learn a stochastic controller for this system
that implements an optimal (or near-optimal) policy. Given a parameter
vector $\theta = (\theta_1,\theta_2, \theta_3, \theta_4)$, we generate
a policy as follows. For any state $x$, let $s_1(x) := \theta_1
\phi_1(x) + \theta_2 \phi_2(x)$, and $s_2(x) := \theta_3 \phi_1(x) +
\theta_4 \phi_2(x)$. 
The probability of choosing action $a_1$ in state $x$ is given by 
$\mu_{a_1}(x) = \frac{e^{s_1(x)}}{ e^{s_1(x)} + e^{s_2(x)}}$, with the 
probability of choosing action $a_2$ given by $1 - \mu_{a_1}(x)$. 
The ratios $\frac{\nabla \mu_{a_i}(x)}{\mu_{a_i}(x)}$ needed by
\pomdpg\ are given by,
\begin{align*}
\frac{\nabla \mu_{a_1}(x)}{\mu_{a_1}(x)}  &= 
\frac{e^{s_2(x)}}{e^{s_1(x)} + e^{s_2(x)}}\left[\phi_1(x), \phi_2(x),
-\phi_1(x), -\phi_2(x)\right] \\
\frac{\nabla \mu_{a_2}(x)}{\mu_{a_2}(x)}  &= 
\frac{e^{s_1(x)}}{e^{s_1(x)} + e^{s_2(x)}}\left[-\phi_1(x), -\phi_2(x),
\phi_1(x), \phi_2(x)\right]. 
\end{align*}
Note that these controllers satisfy assumption \ref{ass:mubound}. 
\vspace{-3mm}
\subsection{Gradient Estimates}
\begin{figure}
\begin{center}
\includegraphics[scale=0.32]{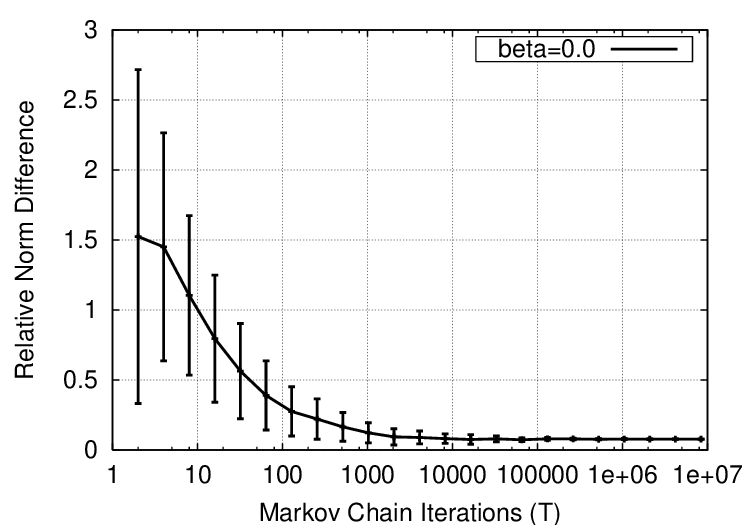}
\includegraphics[scale=0.32]{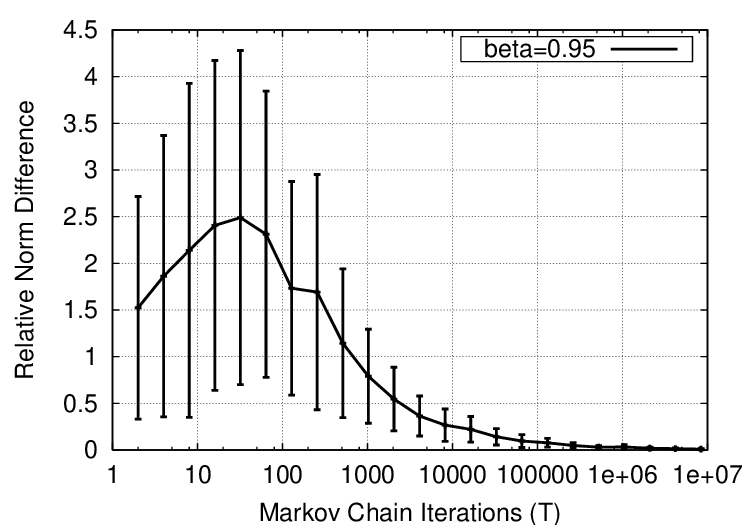}
\caption{A plot of $\frac{\|\nabla \eta - \Delta_T\|}{\|\nabla \eta\|}$
for the three-state Markov chain, for two values of the discount
parameter $\beta$. \label{fig:3statenormsvar}}\vspace{-5mm}
\end{center}
\end{figure}

\begin{figure}
\begin{center}
\includegraphics[scale=0.6]{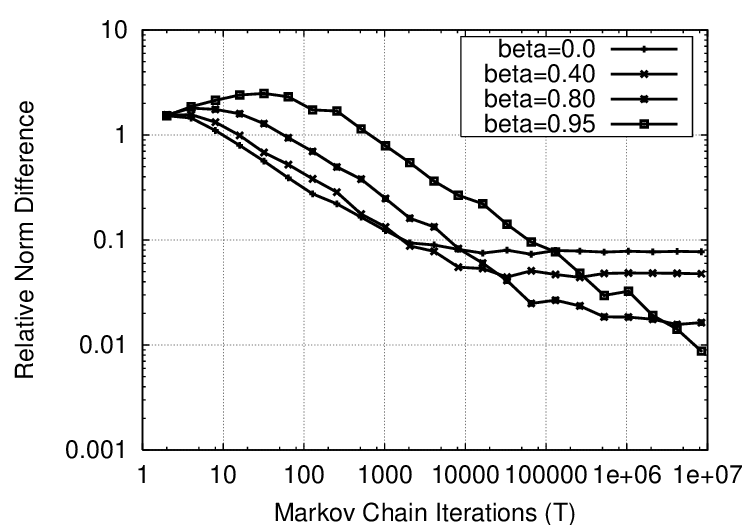}
\caption{Graph showing the final bias in the
estimate $\Delta_T$ (as measured by $\frac{\|\nabla \eta -
\Delta_T\|}{\|\nabla \eta\|}$) as a function of $\beta$ for the
three-state Markov chain. $\Delta_T$ was generated by Algorithm
\ref{algorithm:pgradmdp}. Note both axes are log scales.
\label{fig:3statenormsbias}}\vspace{-5mm}
\end{center}
\end{figure}

\vspace{-2mm}
With a parameter vector\footnote{Other initial values of the parameter
vector were chosen with similar results. Note that $[1,1,-1,-1]$ generates
a suboptimal policy.}
of $\theta = \left[1, 1, -1, -1\right]$, estimates $\Delta_T$ of
$\nb \eta$ were generated using \pomdpg, for various values of $T$
and $\beta\in [0,1)$. To measure the progress of $\Delta_T$ towards
the true gradient $\nabla \eta$, $\nabla \eta$ was calculated from
\eqref{eq:gradeta} and then for each value of $T$ 
the relative error $\frac{\|\Delta_T- \nabla \eta\|}{\|\nabla\eta\|}$
was recorded.  The relative errors are plotted in 
\ref{fig:3statenormsvar} and~\ref{fig:3statenormsbias}. The first
Figure shows how large $\beta$ increases the variance of \pomdpg,
while the second Figure shows a corresponding decrease in the final
bias. Taken together they illustrate nicely the bias/variance
trade-off in the choice of $\beta$. 
\vspace{-3mm}
\subsection{Training via Conjugate-Gradient Ascent}
\label{mdpcg}
\vspace{-2mm}
\sloppy
\conjgrad\ with \pomdpg\ as the ``\grad'' argument was used to train
the parameters of the controller described in the previous
section. Following the low bias observed in the experiments of the
previous section, the argument $\beta$ of \pomdpg\ was set to
$0$. After a small amount of experimentation, the arguments $s_0$ and
$\epsilon$ of \conjgrad\ were set to $100$ and $0.0001$
respectively. None of these values were critical, although the
extremely large initial step-size ($s_0$) did considerably reduce the
time required for the controller to converge to near-optimality. 

\begin{figure}
\begin{center}
\includegraphics[scale=0.6]{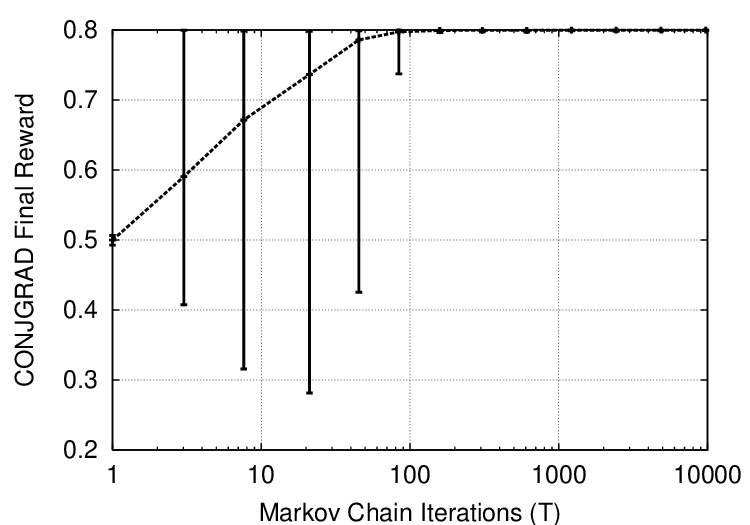}
\caption{Performance of the three-state Markov chain controller trained by 
\conjgrad\ as a function of the total number of iterations of the
Markov chain. 
\label{fig:3statecg}}\vspace{-5mm}
\end{center}
\end{figure}

Figure~\ref{fig:3statecg} shows the average reward $\eta(\theta)$ of
the final controller produced by \conjgrad, as a function of the total
number of simulation steps of the underlying Markov chain.  The plots
represent an average over $500$ independent runs of \conjgrad. Note
that $0.8$ is the average reward of the optimal policy. The parameters
of the controller were (uniformly) randomly initialized in the range
$[-0.1, 0.1]$ before each call to \conjgrad. After each call to
\conjgrad, the average reward of the resulting controller was computed
exactly by calculating the stationary distribution for the
controller. 

\section{Conclusion}
\label{section:conc}
Gradient-based approached to reinforcement learning hold much promise
as a means to solve problems of partial observability and to avoid
some of the pitfalls associated with non-convergence of value-function
methods. A third reason for preferring direct policy approaches is that
it is often far easier to construct a reasonable class of
parameterized policies than it is to construct a class of value
functions; we often know {\em how} to act without being able to
compute the {\em value} of acting. 

In this paper we have analyzed one algorithm for computing an
approximation to the performance gradient. There should be many
possible generalizations to other approximate algorithms. We also
showed how the approximate gradients could be used robustly in greedy
local search. One weakness of our algorithm is the need to specify
running times and the discount factor $\beta$ in advance. we are
currently investigating automatic algorithms for finding these
variables.

It is somewhat ``folklore'' in the Machine Learning community that
gradient-based methods suffer from unacceptably large variance. The
reasons for this conclusion are still not clear and warrant further
investigation.
There are also many avenues for further research. Particularly exciting is
the generalization of \pomdpg\ to multi-agent settings, and
implications for learning in biological neural networks \cite{drl3_99}.

\bibliographystyle{mlapa}
{\small
  \bibliography{icml2k}
  }
\end{document}